\documentclass[11pt,a4paper]{article}
\usepackage[hyperref]{emnlp2020}
\usepackage{times}
\usepackage{latexsym}
\usepackage{booktabs}
\usepackage{graphicx}

\usepackage{color}
\usepackage{bbm}
\usepackage{amsmath}
\usepackage{enumitem}
\usepackage{microtype}

\aclfinalcopy 
\def\aclpaperid{8} 

\title{Learning Adaptive Language Interfaces through Decomposition}

\author{Siddharth Karamcheti \and Dorsa Sadigh \and Percy Liang \\
        Computer Science Department, Stanford University \\
        \texttt{\{skaramcheti, pliang, dorsa\}@cs.stanford.edu} 
}

\date{}

\newcommand{\notsure}{\texttt{NOT-SURE}}

\begin{document}
\maketitle

\begin{abstract}

Our goal is to create an interactive natural language interface that efficiently and reliably learns from users to complete tasks in simulated robotics settings. We introduce a neural semantic parsing system that learns new high-level abstractions through \textit{decomposition}: users interactively teach the system by breaking down high-level utterances describing novel behavior into low-level steps that it can understand. Unfortunately, existing methods either rely on grammars which parse sentences with limited flexibility, or neural sequence-to-sequence models that do not learn efficiently or reliably from individual examples. Our approach bridges this gap, demonstrating the flexibility of modern neural systems, as well as the one-shot reliable generalization of grammar-based methods. Our crowdsourced interactive experiments suggest that over time, users complete complex tasks more efficiently while using our system by leveraging what they just taught. At the same time, getting users to trust the system enough to be incentivized to teach high-level utterances is still an ongoing challenge. We end with a discussion of some of the obstacles we need to overcome to fully realize the potential of the interactive paradigm.

\end{abstract}

\section{Introduction}

\begin{figure}[t]
    \centering
    \includegraphics[width=\linewidth]{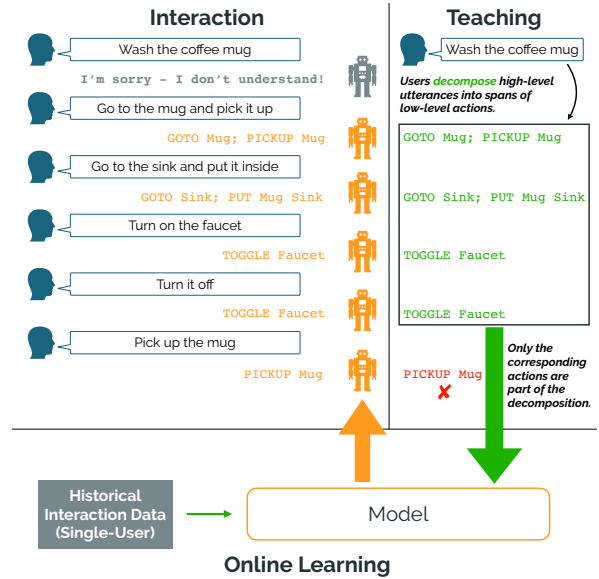}
    \caption{In our proposed framework, users interact with a simulated robot to complete tasks. Central to our approach is \textit{learning by decomposition}: users teach the system to understand novel high-level utterances by breaking them down into utterances that the system can understand and execute. Using these decompositions, we update a semantic parser online, allowing our system to adapt to users as they complete more tasks.}
    \label{fig:framework}
\end{figure}

\begin{figure*}[t!]
    \centering
    \includegraphics[width=\linewidth]{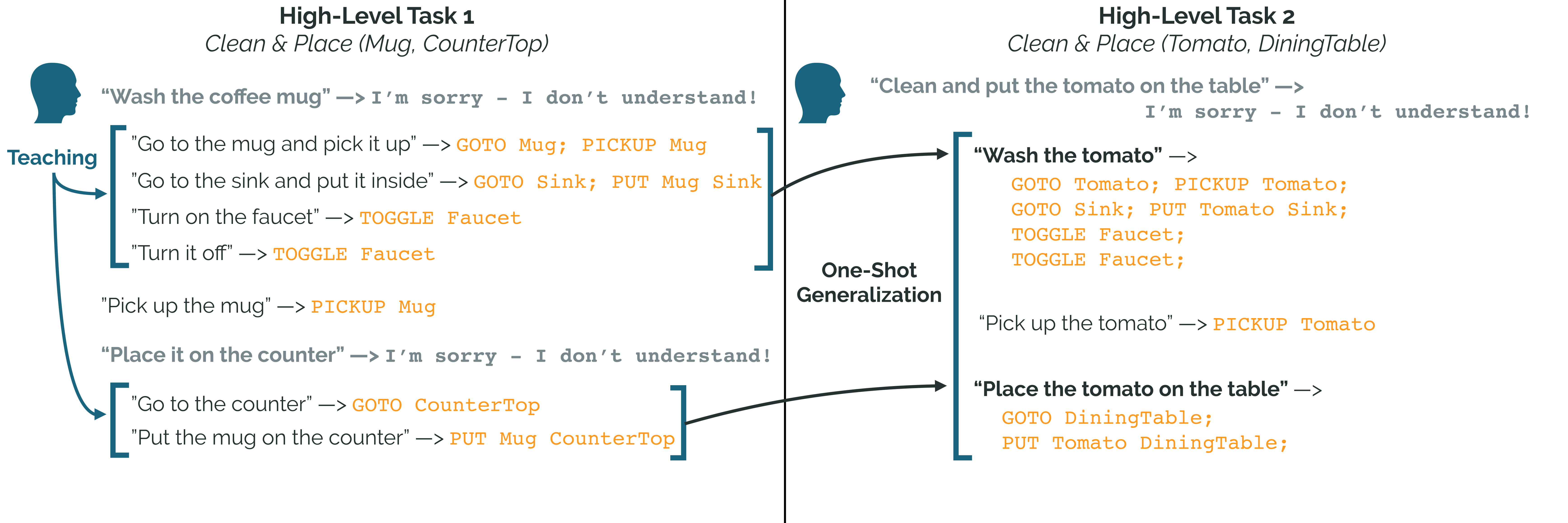}
    \caption{One-shot generalization example: When the system fails to understand an utterance (e.g. ``wash the coffee mug'', ``place it on the counter''), the user teaches the system by \textit{decomposing} it into other utterances the system can understand (illustrated by brackets above), which eventually get mapped to low-level actions that are executed. This induced mapping of high-level utterance to low-level actions forms an example that we use to update our semantic parser online. Because our semantic parser is capable of reliable \textit{one-shot generalization}, users can leverage these decompositions when completing the next task.}
    \label{fig:front}
\end{figure*}

As robots are deployed in collaborative applications like healthcare and household assistance \citep{scassellati2012robots, knepper2013ikeabot}, there is a growing need for reliable human-robot communication. One such communication modality that is both user-friendly and versatile is natural language; to this end, we focus on robust natural language interfaces (NLIs) that can map utterances to executable behavior \citep{tellex2011understanding, artzi2013weakly, thomason2015learning, arumugam2017accurately, shridhar2020alfred}.

Most existing work on NLIs (and AI systems more broadly) falls into a static train-then-deploy paradigm: models are first trained on large datasets of (language, action) pairs and then deployed, with the hope they will reliably generalize to new utterances. Yet, what happens when such models make mistakes or are faced with types of utterances unseen at training --- for example, providing a household robot with a novel utterance like ``wash the coffee mug?'' Such static systems will fail with no way to recover, burdening the user to find alternate utterances to accomplish the task (or give up). Instead, we argue that NLIs need to be dynamic and adaptive, learning interactively from user feedback to index and perform more complicated behaviors. 

In this work, we explore building NLIs for simulated robotics that learn from real humans. Inspired by \citet{wang2017naturalizing}, we leverage the idea of \textit{learning from decomposition} to learn new abstractions. Just like how a human interactively teaches a new task to a friend by breaking it down, users interactively teach our system by simplifying utterances that the system cannot understand (e.g. ``wash the coffee mug'') into lower-level utterances that it can (e.g. ``go to the coffee mug and pick it up'', ``go to the sink and put it inside'', etc. --- see Figure \ref{fig:framework}).

To map language to executable behavior, \citet{wang2017naturalizing} and \citet{thomason2019improving} built adaptive NLIs that leverage grammar-based parsers that allow reliable \textit{one-shot generalization} but lack \textit{lexical flexibility}. For example, a grammar-based system that understands how to ``wash the coffee mug'' may not generalize to ``clean the mug.'' Meanwhile, recent semantic parsers are based primarily on neural sequence-to-sequence models \citep{dong2016logical, jia2016recombination, guu2017bridging}. While these models excel from a \textit{lexical flexibility} perspective, they lack the ability to perform reliable \textit{one-shot generalization}: it is difficult to train them to generalize from individual examples \citep{koehn2017six}.  

In this paper we propose a new interactive NLI that is \textit{lexically flexible} and can \textit{reliably and efficiently perform one-shot generalization}. We introduce a novel exemplar-based neural network semantic parser that first abstracts away entities (e.g. ``wash the coffee mug'' $\rightarrow$ ``wash the $<$obj$>$''), allowing for generalization to previously taught utterances with novel object combinations. Our parser then retrieves the corresponding ``lifted'' utterance and respective program (exemplar) from the training examples based on a learned metric (implemented as a neural network), giving us the lexical flexibility of sequence-to-sequence models.

We demonstrate the efficacy of our learning from decomposition framework through a set of human-in-the-loop experiments where crowdworkers use our NLI to solve a suite of simulated robotics tasks in household environments. Crucially, after completing a task, we update the semantic parser so that users can immediately reuse what they taught. We show that over time, users are able to complete complex tasks (requiring several steps) more efficiently with our exemplar-based method compared to a neural sequence-to-sequence baseline. However, for more straightforward tasks that can be completed in fewer steps, we see similar performance to the baseline. We end with an error analysis and discussion of user trust and incentives in the context of building interactive semantic parsing systems, paving the way for future work that better realizes the potential of the interactive paradigm.

\section{Learning from Decomposition}

User sessions are broken up into a sequence of \textit{episodes} (individual tasks), each comprised of \textit{two phases}: 1) \textit{Interaction}, where the user provides utterances to the system to accomplish the task, and 2) \textit{Teaching}, where the user teaches the system to understand novel utterances (Figures \ref{fig:framework} and \ref{fig:front}).

\subsection{Interaction}

\begin{table}[t]
\centering
\begin{tabular}{@{}ll@{}}
\toprule
Primitive Action         & Canonical Utterance \\ \midrule
\texttt{GOTO <OBJ>}       & go to $<$obj$>$        \\
\texttt{PICKUP <OBJ> }    & pick up $<$obj$>$      \\
\texttt{OPEN <OBJ> }      & open $<$obj$>$         \\
\texttt{CLOSE <OBJ> }     & close $<$obj$>$        \\
\texttt{TOGGLE <OBJ> }    & turn on/off $<$obj$>$  \\
\texttt{PUT <OBJ> <OBJ> } & put object $<$obj$>$   \\ \bottomrule
\end{tabular}
\caption{List of primitive programmatic actions and seed utterances used to initialize our semantic parser. Note that the utterances are \textit{lifted}; they do not include references to concrete objects. This enables one-shot generalization to unseen object combinations.}
\label{table:primitives}
\end{table}

During interaction, the user attempts to complete a task by producing a sequence of user utterances $u_1, u_2, \ldots$ with the corresponding system responses $p_1, p_2, \ldots$ (including the \notsure{} action) that are executed in the environment (the \notsure{} action executes to an error message ``I'm sorry - I don't understand!''). For example, in Figure \ref{fig:framework}, the user first says the novel utterance ``wash the coffee mug,'' and the system returns \notsure{}. The user follows up with ``go to the mug and pick it up,'' which the system maps to the program \texttt{GOTO Mug; PICKUP Mug}. This continues until the user has completed the task. If the system or user makes a mistake and produces an undesired action, the user must continue to provide utterances, as there are no resets.

\subsection{Teaching}
\label{sec:teaching}

The goal of teaching is to convert the sequence of utterance-action pairs $(u_i, p_i)$ into a set of valid training examples for updating the system. To do this, the system presents the user with each $u_i$ where $p_i$ is \notsure{}, and asks the user to select the corresponding contiguous sequence of actions $p_{i + 1}, \ldots p_j$. To facilitate comprehension, we show users (programatically generated) human-readable representations of each action $p$ --- e.g. ``go to the mug'' for a program $p = $ \texttt{GOTO Mug}. For example, the user maps ``wash the coffee mug'' to the sequence \texttt{GOTO Mug; PICKUP Mug; $\ldots$ TOGGLE Faucet} (see Figure \ref{fig:framework} for the full decomposition). Similarly, the user maps ``place it on the counter'' to \texttt{GOTO CounterTop; PUT Mug CounterTop}. The resulting examples $(u_i, \hat{p}_i = p_{i + 1} \ldots p_j)$ are used to update the system (details in Section \ref{sec:semparse-update}). We update \textit{every time} a user completes a task and teaches new examples --- this allows users to access what they have taught immediately, during the following task.

\subsection{Desiderata}

This example illustrates two desiderata for our framework, both of which are key to \textit{trust}: 1) the ability to identify novel types of utterances (when to output \notsure{}), as well as 2) the ability to perform one-shot generalization. Knowing when to output \notsure{} is key to \textit{trust during inference}: signaling to users what the system knows, so that the simulated robot does not take undesired actions (like dropping your coffee mug on the floor). Performing one-shot generalization is key to \textit{trust during learning}: users need to rely on the system remembering what has been taught so they can more efficiently complete future tasks. For example, when the user is completing the next task (second half of Figure \ref{fig:framework}), they should be able to rely on the system understanding ``wash the tomato'' and ``place the tomato on the table,'' even though these refer to different objects than in the taught examples. Section \ref{sec:int-semparse} discusses how we enable one-shot generalization in further detail.

\paragraph{Sequence-to-sequence models fail.}

We found modern neural sequence-to-sequence models to be a poor fit in our setting. The biggest problem we found was their ability to handle novel utterances. Anecdotally, we found when given the novel utterance ``wash the coffee mug,'' a neural sequence-to-sequence system trained on the seed set of utterances in Table \ref{table:primitives} returned the program \texttt{OPEN Mug}, which does not even execute. These problems are exacerbated by the lack of training data; a single user's interaction only creates a handful of new examples, contraindicating the use of data-hungry sequence-to-sequence models \citep{koehn2017six}.

\section{Semantic Parsing}
\label{sec:int-semparse}

To address the above desiderata (identifying when to output \notsure{}, and one-shot generalization), we incorporate two key insights into our approach. To identify when to output \notsure{}, we look at the distances between a new utterance and the utterances in our training set, similar to the exemplar-based approach of \citet{papernot2018deep} --- if an utterance is ``close enough'' to a training utterance, return the corresponding program, otherwise return \notsure{}. To enable one-shot generalization, our parser operates over \textit{lifted} versions of utterances and programs --- versions that abstract out explicit references to objects (allowing for automatic generalization to new combinations of objects unseen during training).

We now describe our semantic parser, which maps a user utterance $u$ and environment state $s$ to the corresponding program $p$ that best reflects the meaning of the user's utterance. In this work, a state $s$ consists of a set of objects where each object is defined by a fixed set of features (e.g. \textit{visibility}, \textit{toggle status}, etc.). We define a program $p$ as a sequence of primitive actions, where each action consists of a template (from Table \ref{table:primitives}) with arguments corresponding to object types. We conclude with a description of how we retrain our semantic parser using the newly taught examples from the teaching phase (Section \ref{sec:teaching}).

\subsection{Model}

\begin{figure}[t]
    \centering
    \includegraphics[width=\linewidth]{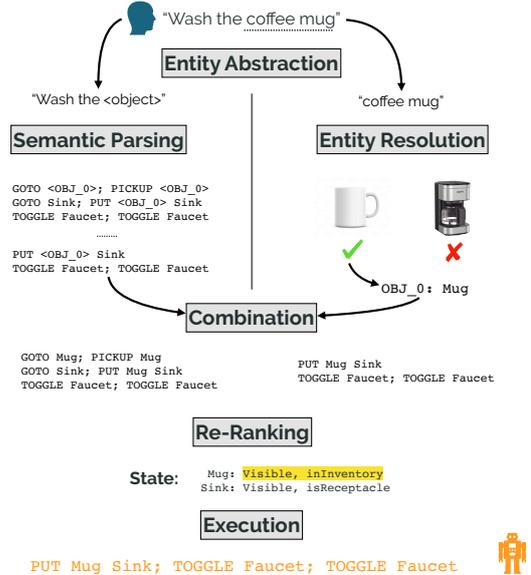}
    \caption{Semantic parsing pipeline. First, entities are extracted and the corresponding outputs --- the lifted utterance and object references --- are parsed into programs and grounded object types. These are combined and re-ranked to identify the program to execute.}
    \label{fig:pipeline}
\end{figure}

Our semantic parser (Figure \ref{fig:pipeline}) takes an utterance $u$ and first abstracts out entities (Section \ref{sec:ent-abs-res}), creating \textit{object references} and \textit{lifted utterances}. We parse these into \textit{object types} (Section \ref{sec:ent-abs-res}) and \textit{lifted programs} (Section \ref{sec:semparse}), which are combined (Section \ref{sec:combination}) and fed to a reranker that additionally uses the state $s$ (Section \ref{sec:reranker}) to identify the program $p^*$ to execute.

\subsubsection{Entity Abstraction \& Resolution}
\label{sec:ent-abs-res}

We define an entity abstractor that maps an utterance $u$ (e.g. ``wash the coffee mug'') to a lifted utterance $f$ (e.g. ``wash the $<$obj$>$'') and a list of object references $\mathcal{O}$ (e.g. [``coffee mug'']). The entity resolver maps each object reference $o \in \mathcal{O}$ (e.g. ``coffee mug'') to a grounded object type $g$ (e.g. \texttt{Mug}) resulting in a new list $\mathcal{G}$. To do this, we exploit a set of ``typical names,'' (e.g. \texttt{Mug} = \{``coffee mug'', ``mug'', ``cup''\}) that we define a priori, looking up the object type with the given name. However, if there are multiple types that share the given name (e.g. in our dataset, table is a ``typical name'' for \texttt{DiningTable, CoffeeTable, SideTable}), we use the current state $s$ to disambiguate: we fetch all the matching items in $s$ and return the physically closest one.

\subsubsection{Semantic Parsing}
\label{sec:semparse}

Central to our approach is the exemplar-based semantic parser that maps a lifted utterance $f$ to a set of lifted programs $\mathcal{Q}$. To do this, we learn a classifier $p_\theta$ that takes two lifted utterances $(f, f')$ and predicts a probability whether they have the same lifted program ($q = q'$). We take $\mathcal{Q}$ to be the programs corresponding to the highest probability $f'$ under $p_\theta$.

\paragraph{Embedding Utterances.}
We first embed each utterance with an embedding function $\phi$, implemented as a neural network that first uses GloVe \citep{pennington2014glove} to embed the words in $f$ followed by position encoding similar to that used in \citet{vaswani2017attention} and a nonlinear transform. The resulting embeddings are summed and fed into to a two-layer MLP to create the utterance embedding $\phi(f)$. The classifier $p_\theta$ outputs $\sigma(a \text{ cos-sim}(\phi(f), \phi(f')) + b)$, where $\text{cos-sim}$ is cosine similarity, $a, b$ are learned scalars, and $\sigma$ is the sigmoid function. We train $p_\theta$ with a binary cross-entropy objective on a training set of (lifted utterance, lifted program) pairs: $\{ (f_i, f_j, [q_i = q_j]) : i, j \in [n] \}$.

\paragraph{Efficient Inference.}
We now describe how we use $p_\theta$ for inference given a new lifted utterance $f'$. Unfortunately, na\"ive application of $p_\theta$ for a new $f'$ requires pairwise comparison with \textit{every training example}. We streamline this by using the structure of our embedding space --- as the classifier outputs the scaled cosine similarity between two utterances, we store the embeddings $\phi(f_i)$ for each training utterance $(f_i, q_i)$ in our dataset, then use an approximate nearest neighbors algorithm to find the the set of utterances that are ``close-enough''; we use the corresponding lifted programs to form the output set $\mathcal{Q}$. We formalize what it means for an utterance to be ``close-enough'' in the following paragraph. We note that this procedure is similar to \textsc{CosineBERT} \citep{mussman2020pairwise}, a model used for active learning on pairwise language tasks.

\paragraph{Setting a Threshold.}
\label{sec:set-threshold}

One of the desiderata of our system is returning \notsure{} for utterances it is not confident about. To do this, we set a \textit{threshold} $\tau$ such that if $\|\phi(f) - \phi(f')\|_2 \ge \tau$, return \notsure{}. Note that this is equivalent to to thresholding the probability output by $p_\theta$ which is monotonic in the cosine distance as defined above. We set this threshold using a held-out validation set of (utterance, program) pairs (defined based solely on the seed examples in Table \ref{table:primitives}). For each utterance in the validation set $f$, we set $\tau$ such that 90\% of the programs corresponding to utterances with $\tau$ are correct. Given an utterance $f'$ at test time, we return the set of lifted programs $\mathcal{Q}$ corresponding to all lifted utterances within $\tau$ of $\phi(f')$ (all lifted utterances ``close enough'' to $f'$).

\paragraph{Handling Compositionality.}

For multi-action utterances (e.g. ``go to the apple \textbf{and} pick it up'') we heuristically split on the keyword ``and,'' resulting in multiple substrings. We parse each substring obtaining subsets of lifted programs, and take the cross-product of these subsets as the final set $\mathcal{Q}$. We acknowledge that this is not a perfect heuristic; in future work we hope to explore more general extensions that allow us to efficiently interpret utterances that have been composed in this way.

\paragraph{Implementation Details.}

When identifying the threshold $\tau$, we define a hyperparameter lower bound $\beta$; this lower bound ensures that our semantic parser isn't overly conservative (returning \notsure{} despite being moderately confident about the set of candidate programs). We find a value $\beta = 0.15$ works well for our experiments. We use Spotify's \texttt{annoy} library as our approximate nearest neighbors store for fast lookups.

We initialize our exemplar-based parser with seed examples (utterances mapped to programs) that cover the set of actions. Table \ref{table:primitives} shows these actions, and a subset of the utterances used for training --- our full dataset consists of only 44 examples (minor variations of the trigger words in the table). This is similar to prior work that defines a set of \textit{canonical utterances} \citep{wang2015overnight}, or a \textit{core grammar} \citep{wang2017naturalizing}. We strip stop words (\textit{the, up, down, on, off, of, in, to, then, a, an, back, front, out, from, with, inside, outside, below, above, top}) from $f$ prior to feeding to our parser to make our model more robust to minor lexical variation. 

\subsubsection{Combination}
\label{sec:combination}

We combine each lifted program $q \in \mathcal{Q}$ with the grounded object types $\mathcal{G}$ to form a set of grounded programs $\mathcal{P} = \{p_1, \ldots, p_k\}$. In general, given a lifted program $q$ that takes a sequence of arguments (e.g \texttt{PUT <OBJ> <OBJ>}) and a list of object types (e.g. $\mathcal{G} = [\texttt{Mug, DiningTable}]$), we simply substitute the object types into the program, replacing each argument in the lifted program. This results in a final grounded program (e.g. $p = \texttt{PUT Mug DiningTable}$).

\subsubsection{Reranking}
\label{sec:reranker}

The semantic parser, entity resolver, and combination step produce a set of grounded programs $\mathcal{P}$. The reranker takes the original utterance $u$, current state $s$, and this set of grounded programs $\mathcal{P}$ and chooses a single candidate $p^* \in \mathcal{P}$ to execute.

As a first step, we discard candidate programs that fail to execute in our simulator: for example, \texttt{PICKUP Mug} is discarded if the robot is already holding an object. Then we use a neural network to produce a score for each $p_i \in \mathcal{P}$. This network separately embeds the utterance, state, and each candidate program, feeding the concatenated embeddings to a two-layer MLP to produce a real-valued score for each $p_i$. In our work, the state $s$ is retrieved dynamically based on the grounded objects $\mathcal{G}$ returned by the entity resolver; the state is made up of hand-coded features corresponding to attributes like \textit{visibility}, \textit{toggle status}, and \textit{whether it can be picked up}, amongst others. We use a similar scheme as the semantic parser (Section \ref{sec:semparse}) to encode utterances and candidate programs (embed, position encode, and sum), and a simple linear transformation to encode the bag-of-features representing the state $s$.

The highest-scoring candidate $p^* \in \mathcal{P}$ is executed. The reranker is trained via the process described in Section \ref{sec:ranker-update} \textit{only after new examples are taught by users} during the teaching phase following each task they are asked to complete.

\subsection{Retraining from User Feedback}
\label{sec:retraining}

In the following subsections, we discuss how to retrain our  semantic parser and reranker to achieve the second of the two desiderata desired of our system: reliable and efficient \textit{one-shot generalization}. As input to the retraining procedure, we take the dataset $\mathcal{\hat{D}} = (u_i, \hat{p}_i)$ of newly taught examples from the teaching phase (Section \ref{sec:teaching}).

\subsubsection{Creating Lifted Examples}
\label{sec:abstract-examples}

Retraining the exemplar-based semantic parser requires converting our grounded dataset $\mathcal{\hat{D}}$ to pairs of lifted utterances and programs. Consider the grounded example (``Place the tomato on the table'', \texttt{GOTO DiningTable; PUT Tomato DiningTable}); we want to map this to its lifted form (``Place the $<$obj$>$ on the $<$obj$>$'', \texttt{GOTO $<$OBJ$>$ PUT $<$OBJ$>$ $<$OBJ$>$}). To do this, we use the entity abstractor and resolver (from Section \ref{sec:ent-abs-res}) to factor out object references. 

Concretely, using the entity abstractor on the above example leaves us with $\hat{f} =$ ``Place the $<$obj$>$ on the $<$obj$>$'', and references $\mathcal{\hat{O}} = $ [``tomato'', ``dining table''], which the entity resolver maps to $\mathcal{\hat{G}} = $ [\texttt{Tomato, DiningTable}]. We replace any element of $\mathcal{G}$ that occurs in the original program with the generic \texttt{$<$OBJ$>$} token to create the lifted program ($\hat{q} = $ \texttt{GOTO $<$OBJ$>$; PUT $<$OBJ$>$}). Applying this procedure to each example in $\mathcal{\hat{D}}$ gives us our lifted examples ($\hat{f}, \hat{q}$).

\subsubsection{Updating the Semantic Parser}
\label{sec:semparse-update}

Updating the semantic parser requires optimizing the binary cross-entropy objective from Section \ref{sec:semparse} using these lifted examples ($\hat{f}, \hat{q}$). As we train our parser from pairs of examples, and there are far more negative examples (pairs with different programs) than positives, we over-sample positive examples so that batches have an equal number of positives and negatives. We train on the entire history of data for the given user, re-creating the nearest neighbors store with embeddings of each training utterance $f_i$. After this step, we re-calibrate the nearest neighbors threshold using the procedure in Section \ref{sec:set-threshold}.

\subsubsection{Updating the Reranker}
\label{sec:ranker-update}
After updating the semantic parser, we re-parse each utterance in our dataset to define our retraining dataset of $(\hat{u}_i, \mathcal{\hat{P}}_i, \hat{s}_i)$ tuples. We use the program $p^*$ that was actually executed for utterance $\hat{u}_i$ in state $\hat{s}_i$ as the ``gold'' label for the reranker. We train the reranker by maximizing the log-likelihood (minimizing the cross-entropy loss) of this candidate $p^*$ amongst the others.

\section{Experiments}

\begin{figure*}[t!]
    \includegraphics[width=.95\linewidth]{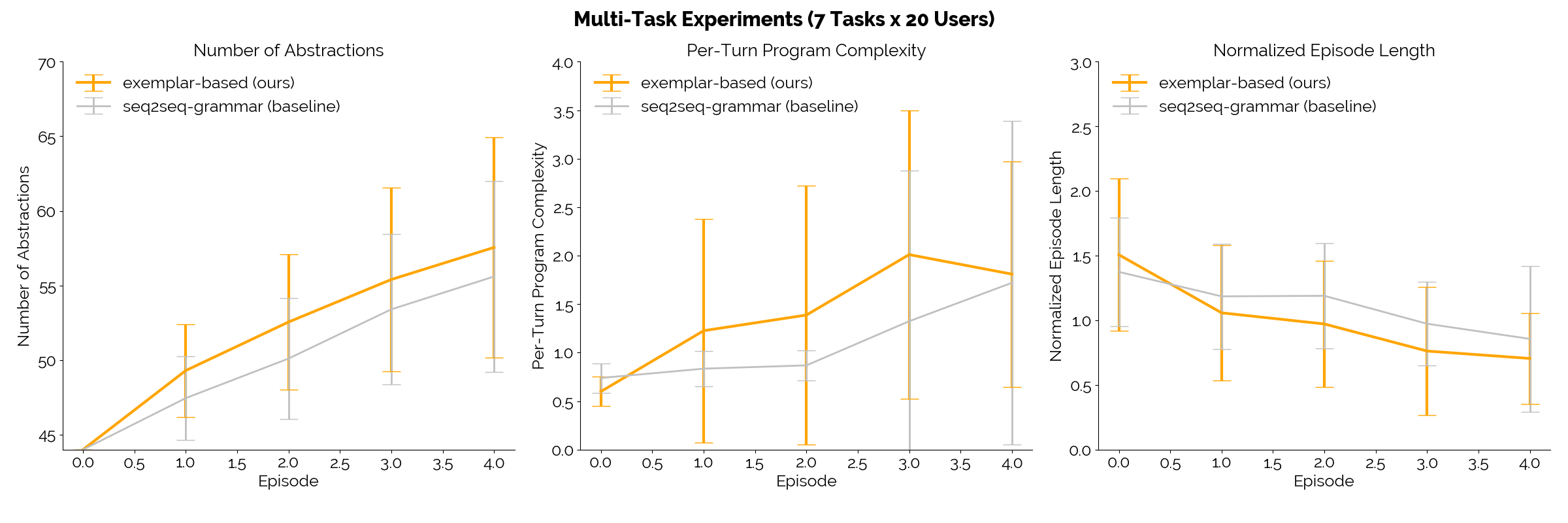}
    \caption{Complete set of results across 20 users with 7 different task types. Each user is given a single task type, and asked to complete 5 different episodes, with different combinations of environments and objects. The graph on the left shows the \textit{number of examples taught} over 5 episodes. The graph in the middle shows the \textit{per-turn program complexity} (number of primitives per language utterance) over time. The last graph shows the \textit{normalized episode length} (\# utterances to solve task / number of actions required).}
    \label{fig:general-results}
\end{figure*}

We evaluate our approach with a set of human-in-the-loop experiments where crowdworkers are tasked with solving a series of simulated robotics tasks. Users interact with our system over 5 episodes (where each episode consists of a single task), teaching our system new examples after successfully completing each one. Each user has their own individual semantic parser and re-ranker (models are not shared across the users), with both components updating online after each teaching phase, prior to the start of the next task. Updating the two models (including rebuilding the nearest neighbors store) after each teaching phase varies depending on task complexity, but takes anywhere from 28 -- 63 seconds on an Amazon EC2 T2.Medium (2 CPUs, 4 GiB RAM, no GPU) instance.

\subsection{Experimental Setup}

\paragraph{Environment and Tasks.}

Our experiments take place in simulated household environments, with users completing structured, everyday tasks. We create a 2D web-client inspired by the AI2-THOR Simulation Environment~\citep{kolve2017ai2thor} that removes the 3D rendering and spatial layout, but preserves the object types, attributes, and relations. 

We borrow our tasks from the \mbox{ALFRED} Dataset \citep{shridhar2020alfred} that defines 7 task types: 1) \textit{Pick and Place}, 2) \textit{Pick Two Objects and Place}, 3) \textit{Look at Object in Light}, 4) \textit{Nested Pick and Place}, 5) \textit{Pick, Clean, and Place}, 6) \textit{Pick, Heat, and Place}, and 7) \textit{Pick, Cool, and Place}.

\paragraph{Interactive User Studies:}

We run our interactive user studies via Amazon Mechanical Turk (AMT). Each user is assigned one of the 7 task types, and is asked to complete 5 tasks of that type in a row. We recruited 20 workers per approach. Workers were paid $\$5$ with an average completion time of 23 minutes. We limit our AMT studies to workers with an approval rating $\geq 98\%$, location = US, and a total number of completed HITs $> 5000$.

\paragraph{Baseline.}
We compare our approach with a neural sequence-to-sequence with attention model similar to \citet{jia2016recombination}. To improve reliability, if the user enters an utterance that can be handled by a simple grammar that covers the core utterances from Table \ref{table:primitives}, we return the resulting program; otherwise, we invoke the sequence-to-sequence model. We find the inclusion of such a grammar necessary to prevent users from getting stuck. We refer to this combination of a neural sequence-to-sequence with a grammar as ``\textit{seq2seq-grammar}'', whereas we refer to our system as ``\textit{exemplar-based}''. We keep the learning by decomposition framework identical for both our system and the sequence-to-sequence system --- in other words, we simply swap out our exemplar-based neural parser described in Section \ref{sec:semparse} for the \textit{seq2seq-grammar} model.

\paragraph{Metrics.}

We define three evaluation metrics: 
\begin{enumerate}[leftmargin=0pt, itemindent=2em]
    \item \textit{Total number of examples taught}: The number of unique (utterance, program) pairs that the users teach the system across each teaching phase (as described in Section \ref{sec:teaching}). This number starts at 44, the number of unique seed examples from Table \ref{table:primitives}. Higher is better --- this metric indicates whether users are engaging with the system to teach high-level abstractions; a flat curve means that the users have finished teaching and are exploiting the examples they have previously taught.
    \item \textit{Per-turn program complexity}: the number of actions generated per utterance. For example, an utterance that generates the program \texttt{GOTO Mug; PICKUP Mug; GOTO Sink; PUT Mug Sink} has complexity of 4 --- one for each primitive (\notsure{} counts at 0). We expect a steep upward trend in this metric over time as users teach and reuse progressively more complex examples. 
    \item \textit{Normalized episode length}: the number of language utterances the user provided divided by the number of primitive actions required to solve the task. This is the end-to-end metric we seek to optimize --- values less than 1 indicate that users are able to tap into what they have taught to complete tasks in fewer steps.
\end{enumerate}

\subsection{Results}
\label{sec:results}

\begin{figure*}[t]
    \includegraphics[width=.95\linewidth]{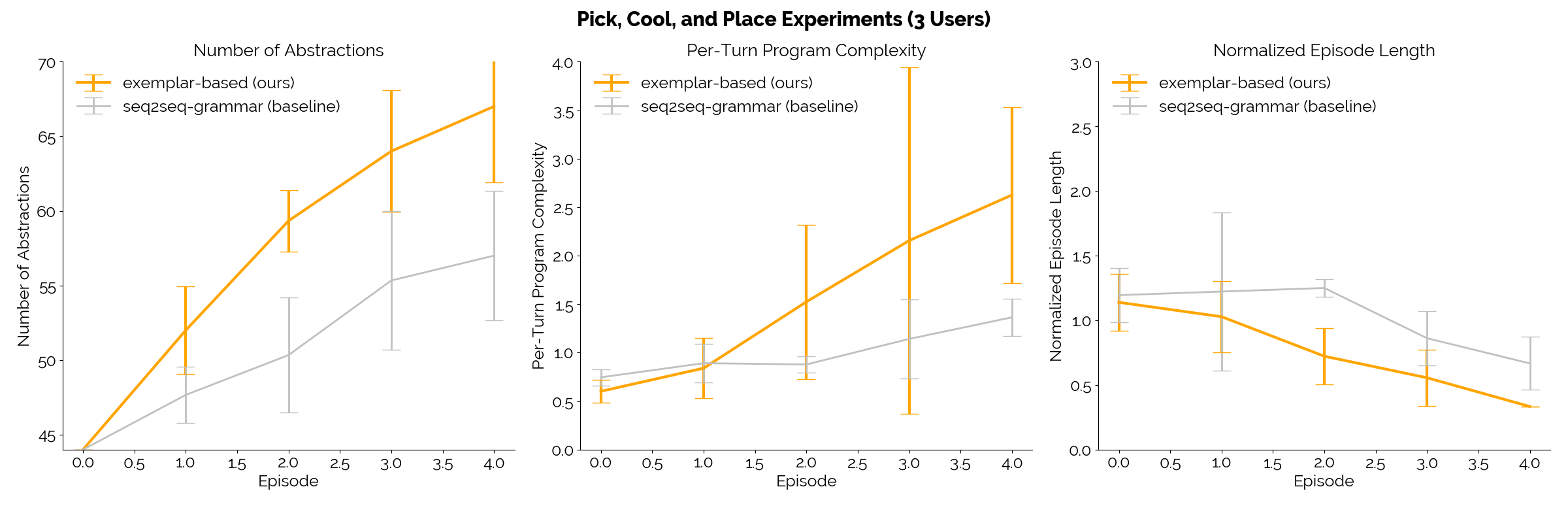}
    \caption{Results for the \textit{Pick, Cool, and Place} task across 3 users (subset of the original 20). This task is complex, requiring at least 12 primitives to complete. Notice how the number of defined examples and per-turn program complexity are much higher for our method, and that the normalized episode length is lower.}
    \label{fig:cool-place}
\end{figure*}

\paragraph{Full Results: 20 Users x 7 Tasks.}

Figure \ref{fig:general-results} presents graphs of the three metrics over the 5 episodes for each of the 20 users, split across the 7 different task. Error bars denote estimated standard deviation across all 20 users. Users of both our exemplar-based system and the sequence-to-sequence baseline teach a moderate number of new examples over time, with an upwards trend in per-turn program complexity as they complete more tasks. Finally, we see a decreasing trend in the normalized episode length, with the mean value of our system dipping slightly below a value of 1 after completing 5 instances.

\paragraph{Case Study: \textit{Pick, Cool, and Place}.}

Figure \ref{fig:cool-place}, on the other hand, presents graphs of the 3 metrics across 3 users for the \textit{Pick, Cool, and Place} task, one of the more complex tasks in our suite, requiring at least 12 primitive utterances to complete successfully (compared to tasks like \textit{Pick and Place} that only require 4). Here we see large gaps between our system and the sequence-to-sequence baseline --- not only do users of our system teach significantly more high-level examples, but they have a much-higher per-turn program complexity after 5 episodes compared to the baseline. Finally, we see that after 5 episodes, the normalized episode length is around 0.2, indicating that users are able to complete the complex task in 1/5 the steps necessary with our system.

\paragraph{Are users re-using high-level abstractions?}

The general results in Figure \ref{fig:general-results} indicate that while users are teaching the system new abstractions, they are unfortunately not re-using them effectively. The normalized episode length plot shows that both systems converge to 1, indicating that users are defaulting to the primitive actions, rather than trying to teach higher-level examples. One possible explanation for this is that for simpler tasks (e.g. \textit{Pick and Place}), it is perhaps easier and faster to provide low-level utterances (those in Table \ref{table:primitives}), rather than teach new examples. Defaulting to low-level utterances also explains the lack of a significant gap between the sequence-to-sequence model and our model --- in light of low-level utterances, the grammar does the heavy-lifting (in other words, we would not be invoking the sequence-to-sequence model at all). Indeed, across all 20 users for the \textit{seq2seq-grammar} model, 89.9\% of successfully parsed utterances (713 out of 793 total) were handled by the grammar, with only 10.1\% handled by the seq2seq model (70 of 793 total). 

However, this trend doesn't hold true for more complex tasks. Figure \ref{fig:cool-place} shows that users are teaching and reusing a significant number of examples, completing tasks extremely efficiently. One hypothesis is to correlate task complexity with abstraction reuse (and thus, the ease by which users solve tasks), and while supported by the \textit{Pick, Cool, and Place} results (Figure {\ref{fig:cool-place}}), we would require future experiments with a larger number of users before we can draw meaningful conclusions.

\section{Related Work}

We build on a long tradition of learning semantic parsers for mapping language to executable programs \citep{zelle96geoquery, zettlemoyer05ccg, zettlemoyer07relaxed, liang11dcs}, with a focus on using context and learning from interaction.

\paragraph{Contextual Semantic Parsing.}

In many settings, successfully parsing an utterance requires reasoning about both linguistic and environment context. \citet{artzi2013weakly} developed a model for parsing instructions in the SAIL Navigation dataset~\citep{macmahon2006walk, chen11navigate} that leverages the environment context. Later, \citet{long2016projections} introduced the SCONE Dataset, requiring building models that can reason over both types of context. More recently, \citet{yu2019cosql} introduced the large-scale Conversational Text-to-SQL (CoSQL) dataset that requires jointly reasoning over dialogue history and databases to parse user queries to SQL. We handle both linguistic context and environment context in our work, by decoupling semantic parsing from grounding; our lifted semantic parser handles linguistic context, while our entity resolver and reranker handle environment context.

\paragraph{Learning from Interaction.}

Closest to our work is Voxelurn \citep{wang2017naturalizing}, and its close predecessor SHRDLURN \citep{wang2016games}. Voxelurn defined an open-ended environment where the goal was to build arbitrary voxel structures using language instructions. We take inspiration from its teaching procedure where users decompose high-level utterances into low-level actions in the context of a grammar-based parser. Other work uses alternative modes of interaction to teach new behaviors. \citet{srivastava2017joint} used natural language explanations to teach new concepts. Relatedly, \citet{labutov2018lia} introduced LIA, a programmable personal assistant that learned from user-provided condition-action rules. Furthermore, \citet{weigelt2020programmingfuse} introduce an approach for teaching systems new programmatic functions from language that explicitly reasons about whether utterances contain ``teaching intents,'' a mechanism that is similar to our procedure for returning \notsure{}. Once these ``teaching intents'' have been identified, they are parsed into corresponding code blocks that can then be executed. Other work leverages conversations to learn new concepts, generating queries for users to respond to \citep{artzi11conversations, thomason2019improving}. Notably, \citet{thomason2019improving} used this conversational structure in a robotics setting similar to ours, but focused on learning new percepts, rather than structural abstractions. \citet{yao2019model} defined a similar conversational system for Text-to-SQL models that decides when intervention is needed, and generates a clarification question accordingly. 

\paragraph{General Instruction Following.}
Other work looks at instruction following for robotics tasks outside the semantic parsing paradigm, for example by mapping language directly to sequences of actions \citep{anderson2018vision, fried2018speakerfollower, shridhar2020alfred}, mapping language to representations of reward functions \citep{arumugam2017accurately, karamcheti2017draggns}, or learning language-conditioned policies via reinforcement learning \citep{hermann2017grounded, chaplot2018gated}.

\section{Discussion \& Lessons Learned}

\paragraph{Towards More Complex Settings.}

Our analysis in Section \ref{sec:results} suggests that situating our system in a more complex setting might allow us to truly see the benefits of learning by decomposition. One such setting is Voxelurn \citep{wang2017naturalizing}, with its open-ended tasks that allow for the definition of multiple different high-level abstractions with compositional richness. In contrast, the tasks in this work are linear, with similar sequences of primitives used to accomplish each high-level task.

Future work should use this insight and identify environments that are more complex and open-ended, where users are naturally incentivized to teach the system new abstractions that built atop each other, to facilitate performing more complex behaviors. In robotics, this might translate to building systems for cooking, perhaps taking inspiration from Epic Kitchens~\citep{damen2018kitchens}, where the set of high-level objectives (general recipes to follow, kitchen behaviors to imitate) is much larger, but where individual subtasks (low-level abstractions like slicing a vegetable, stirring a pot) are very common and generalizable. 
Other settings might include open-ended building tasks, either in the real world \citep{knepper2013ikeabot, lee2019ikea}, or in virtual worlds like Minecraft \citep{johnson2016malmo, gray2019craftassist}.

\paragraph{On Trusting Interactive Learning.}

Users have an implicit expectation that after providing just a single example --- say to ``wash the coffee mug'' --- the system will know how to ``wash the tomato'' or even ``clean the plate'' immediately. However, existing machine learning is not built with such extreme data efficiency in mind; especially for harder types of generalization (e.g. to ``clean the plate''), we cannot guarantee learning this in a single step. While in this work we show reliable one-shot generalization across objects in a simplified setting, the real-world is much more complex, and different entities merit different behaviors. For example, consider generalizing from ``wash the spoon'' to ``wash the table''; a system like ours will try to execute the program taught in the first context (going to the sink, placing the object inside, etc.) to the second, leading to complete failure. 

Part of the problem is a \textit{lack of transparency}; after teaching an example, it is hard for a user to understand what the system knows. This impacts trust, and as a result, when the system makes a mistake interpreting a high-level utterance, users back off to using utterances they are confident the system will understand (mirroring our observed results). This suggests future work in building more reliable methods for one-shot generalization and interpretability, providing users with a clear picture of what the model has learned.

\section*{Acknowledgements}

This work was supported by NSF Award Grant no. 2006388 and the Future of Life Institute. S.K. is supported by the Open Philanthropy Project AI Fellowship. We thank Robin Jia, Michael Xie, John Hewitt, and Chris Potts for helpful feedback during the initial stages of this work. We finally thank the anonymous reviewers for their helpful comments.

\bibliographystyle{acl_natbib}
\bibliography{references}

\end{document}